# Enhanced Consumer Feedback Enabler System for Advertisement Boards using Auto Panning Camera

[1]Aditya Ajit Khadilkar
Vellore Institute of Technology,
Vellore, India
ka.aditya1@gmail.com

[2]Godwyn James William
Vellore Institute of Technology,
Vellore, India
jamesgodwynj@gmail.com

[3]Hemprasad Yashwant Patil
School of electronics engineering
Vellore Institute of Technology,
Vellore, India
hemprasadpatil@gmail.com

*Abstract* - **The feedback of consumers who pass by an advertisement board is crucial for the marketing teams of corporate companies. If the emotions of a consumer are analyzed after exposure to the advertisement, it would help to rate the quality of the advertisement. The state-of-the-art emotion analyzers can do this task seamlessly. However, if the consumer moves away from the center of the advertisement board, it becomes difficult for the camera to capture the person with sufficient details. Here, the role of an auto-pan and tilt camera is imminent if a person moves out from the frame limits of the camera. This paper aims to help solve the above issue by panning and tilting the camera by precise amount automatically using facial detection and interpolation algorithms. We propose a method where a camera attached to servo motors can automatically pan and tilt such that the subject is always in the center of the frame. This would be done by facial detection and interpolation of its position with respect to the angle of the camera. The direction of the camera is controlled with the help of a microcontroller, which takes-in the angle values of where the camera needs to move in order to maintain the subject's face in the center. We have designed a system that works on the Arduino platform and can pan and tilt the camera in real-time.**
*Keywords - Computer vision, microcontroller, camera, facial detection, Consumer feedback analysis.*

## I. INTRODUCTİON

This paper is an application of facial localization algorithms realized on microcontrollers—used to control a set of servo motors through a serial interface. Whenever there is a moving subject in surveillance, keeping it in the center of the frame becomes a vital concern to the automatic security systems. In addition, applications such as filming people, filming high-speed objects, require high sensitivity in panning and tilting the camera e.g. rockets, racing cars, etc. or operating in delicate and sensitive environments like endoscopic cameras to track down moving parasites. On the other hand, in cellular research e.g. observing locomotion of a cell for a long period of time requires such automated tracking.

In this paper, object localization has been used to identify the position of the target object in the frame. The algorithm computes the target's distance from the center of the frame and interpolates it with the angles required to tilt the camera in order to get the object in the center of the frame. It then sends these angles to the microcontroller which turns the servo mount on top of which the camera sits.

## II. METHODOLOGY

### i. Abbreviations

All the distances in this paper are measured in px (pixels) as they deal with a digital image and not a physical distance. The angles are measured in degrees.

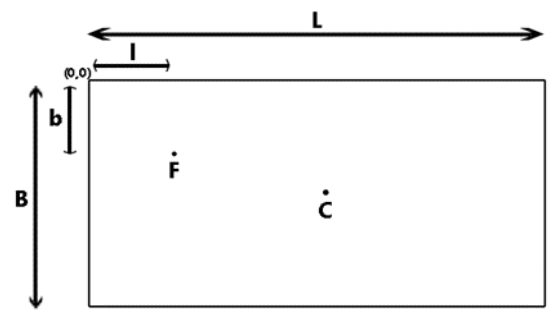

L = Total length of the image in px that the camera can perceive (Horizontal resolution)
B = Total Breadth of the image in px that the camera can perceive (Vertical resolution)
F = Centre point of the face
C = Centre point of the frame
l = X-Coordinate of the point F (pixels)
b = Y-Coordinate of point F (pixels)

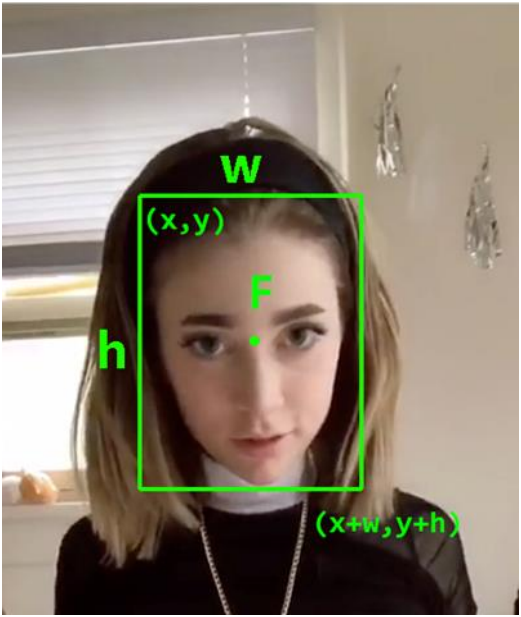

**Fig. 2** Co-ordinate labels for the detected face.

$$l = (x + w/2) \quad (1)$$
$$b = (y + h/2) \quad (2)$$
$$F = (l, b) \quad (3)$$

L/2 and B/2 are where point F should ideally be. i.e. Center of the frame. But it's currently at l,b Hence it needs to move by (L/2) – l and (B/2) – b. We want to change the angle of pan and tilt so that the distance changes. Hence, we will transform change in length to change in angle. This will be done by ratios.

*ii.   Formulation*

δl = change in length = $[(L/2) - l]$
L = max length
θ = maximum horizontal angle range of the camera.
Φ = maximum vertical angle of the camera
δθ = change in theta (angle of panning).

We will be giving δθ as input to the panning servo motor. [YAW]
Since l and θ are proportional to each other;

$$\delta l / L = \delta\theta / \theta$$
$$\delta\theta = (\delta l / L) * \theta$$
$$\delta l = (L/2) - l$$
$$\delta\theta = \{[(L/2) - l]/L\} * \theta$$
$$\delta\theta = [0.5 - (l/L)] * \theta$$
$$\delta\theta = [0.5 - (l/L)] * \theta \quad (4)$$

Similarly, for vertical angle ϕ [PITCH]
$$\delta\varphi = [0.5 - (b/B)] * \varphi \quad (5)$$

δθ and δϕ are the turning angles that we will provide the servo.

The arguments that we will require are current x and y coordinates of the face, Maximum horizontal and vertical resolution of the camera, and the vertical and horizontal angle of the capture of the camera.

*iii.   Notes for replication*

- δθ and δϕ are converted from float to integers. As most servo motors can move only in fixed integer angles (1° least count). If the servo being used has an even lower least count, round up the values accordingly and parse it through the com port as a string.
- The servo used in this project took input from 0° to 180° where 90° was the neutral midpoint. (when δθ was 0° the servo should be 90°)
- δθ and δϕ are in the range (-θ/2 to +θ/2) and (-ϕ/2 to +ϕ/2). Hence, we add 90 to δθ and δϕ. This will shift from 0° to 90°.

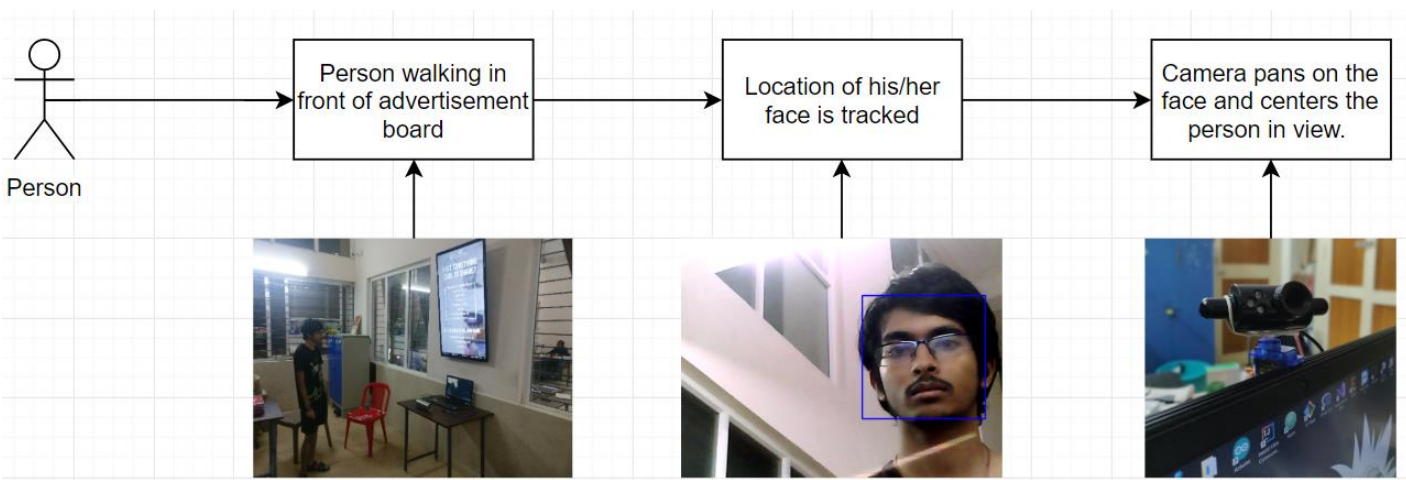

**Fig 3.** Concept and working of the auto panning camera in an advertisement board.

## III. RESULTS AND CONCLUSİONS

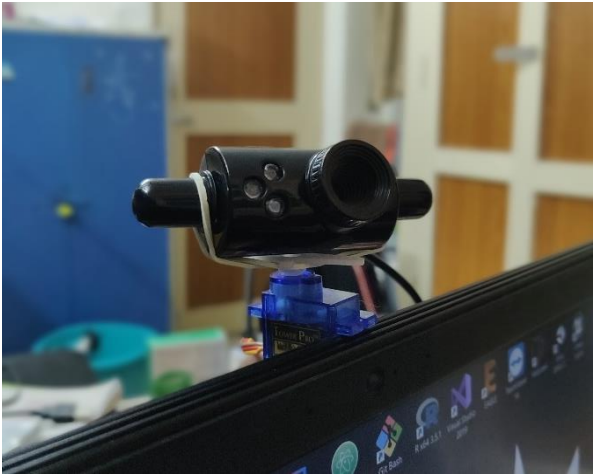

**Fig 4.** Working model of auto panning camera.

We were able to demonstrate the feasibility of a working auto panning camera. The facial data collected by the panning camera can be used in sentiment analysis to identify if the person walking by an advertisement board likes the advertisement or not. This could also help in identifying and using the impact factor of the advertisement being displayed and analyze if the advertisement needs to be improved or not.

This project is generalized and applicable for many other applications apart from advertisement boards. We used the human face as the target object; however, this target can be anything from rockets, aircraft, and automobiles to micro-organisms and parasites and wild animals.
There is ample scope of development on this project, incorporating "Peeking into the future" model to get a pre-motion of where the object will be in the future.